\documentclass[letterpaper,twocolumn,10pt]{article}

\usepackage{hegroup}
\usepackage{graphicx}
\usepackage{float}
\usepackage{amsmath}
\usepackage{array}
\usepackage{xspace}
\usepackage{tabularx}
\usepackage[breakable]{tcolorbox}
\tcbuselibrary{breakable}

\newcommand{\gentext}[1]{{\ttfamily #1}}

\newtcolorbox{promptbox}[1][]{%
  colback=blue!4, colframe=blue!45!cyan!80!black, boxrule=0.6pt, arc=2.5pt,
  left=5pt, right=5pt, top=3pt, bottom=3pt,
  fonttitle=\bfseries\footnotesize, coltitle=white,
  colbacktitle=blue!45!cyan!80!black, title=#1, breakable,
  before skip=6pt, after skip=6pt}

\newcommand{\mypara}[1]{\noindent{\bf {#1}.}\xspace}

\makeatletter
\def\@fnsymbol#1{\ensuremath{\ifcase#1\or \dagger \or \ddagger \or \mathsection \or \mathparagraph \or \| \or ** \or \dagger\dagger \or \ddagger\ddagger \else\@ctrerr\fi}}
\makeatother

\begin{document}
\setcounter{secnumdepth}{3}

\title{IPV-Bench: Benchmarking Image Protection Methods under Diverse Image-to-Video Generation Scenarios}

\author{
Xiaofeng Li\textsuperscript{1}  \ \ \
Leyi Sheng\textsuperscript{1}  \ \ \
Yifan Zhao\textsuperscript{1}  \ \ \
Zhen Sun\textsuperscript{1}  \ \ \
Zongmin Zhang\textsuperscript{1}  \ \ \
Jiaheng Wei\textsuperscript{1}  \ \ \
Xinlei He\textsuperscript{2} \thanks{Corresponding author (\href{mailto:wooohxl@gmail.com}{wooohxl@gmail.com}).} \ \ \
\\
\\
\textsuperscript{1}\textit{The Hong Kong University of Science and Technology (Guangzhou)} \ \ \
\textsuperscript{2}\textit{Wuhan University} \ \ \
\\
}

\date{}

\maketitle

\begin{abstract}
    Image-to-video (I2V) generation models can be misused to animate a single image into a convincing fake video, motivating perturbation-based image protection methods that aim to disrupt such generation.
    Yet these methods remain difficult to compare: they are reported under inconsistent metrics and generation settings, are often validated only on the single generator they were optimized against, and are evaluated on narrow, single-domain image sets that do not reflect real misuse.
    To address these challenges, we introduce \textbf{IPV-Bench} (\textbf{I}mage \textbf{P}rotection against \textbf{V}ideo generation), the first systematic benchmark for image protection in I2V generation scenarios.
    IPV-Bench couples a unified protocol that jointly scores protection effectiveness, visual fidelity, and robustness to preprocessing attacks together with IPV-500, a prompt-paired dataset spanning five misuse-relevant domains. Based on this benchmark, we evaluate five representative protection methods across four I2V models covering distinct architectures and both open-source and commercial systems.
    Extensive experiments show a consistently sobering picture: image protection and video disruption trade off against each other, most methods fail to disrupt generation beyond noise, protection rarely transfers across generators, and the few effective cases are broken by simple preprocessing.
    We further find that image content governs the perceptual cost of protection but not its benefit: no image domain offers an easier target.
    Overall, IPV-Bench provides a rigorous, reproducible, and extensible foundation for developing protection methods that work in practice.
\end{abstract}

\section{Introduction}
\label{sec:intro}

\begin{figure*}[t]
    \centering
    \includegraphics[width=0.95\linewidth]{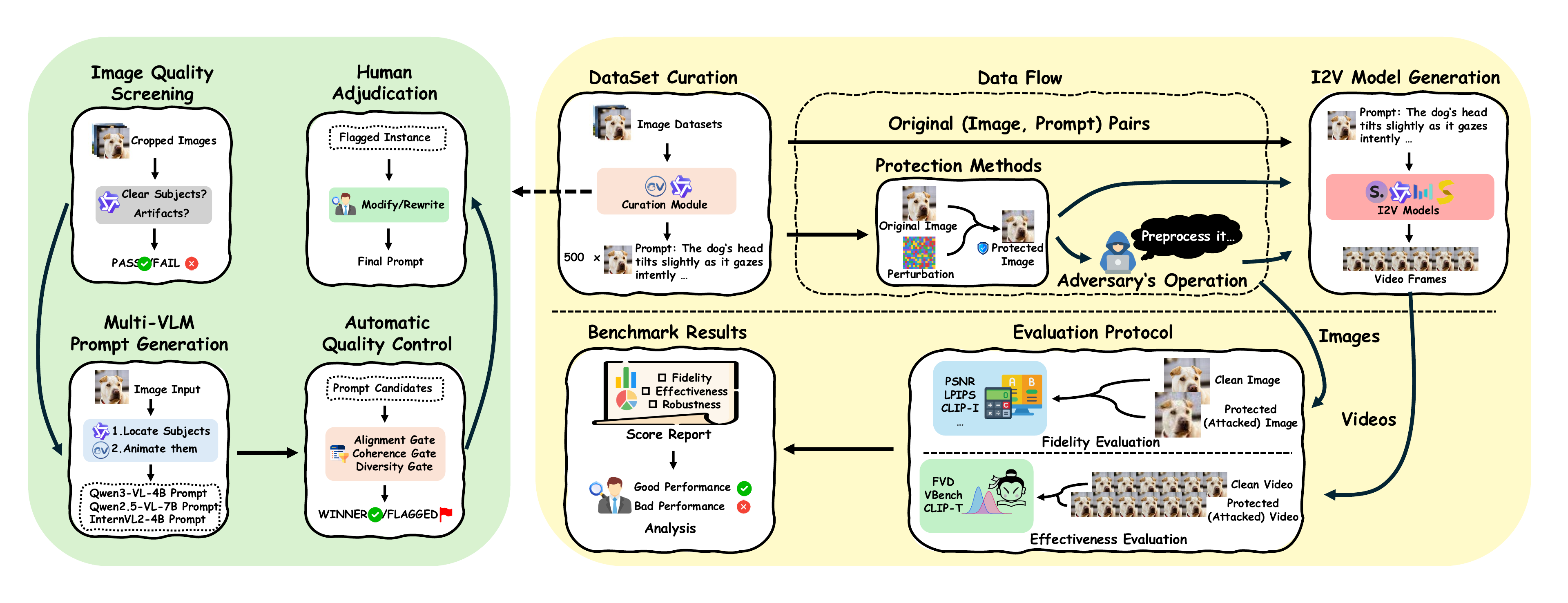}
    \caption{\textbf{Overview of IPV-Bench.} The data-curation pipeline is expanded on the left part of the figure: each image passes a VLM quality-screening gate, is paired with a motion-oriented prompt from multiple VLMs, filtered by automatic quality gates, and finalized by human adjudication, yielding IPV-500. The right part shows the evaluation flow: a protection method perturbs each image, an adversary may preprocess it, and I2V models generate videos, which are scored on fidelity, effectiveness, and robustness.}
    \label{fig:mainfigure}
\end{figure*}

Image-to-Video (I2V) generation has experienced rapid progress in recent years~\cite{dhariwal2021diffusionmodelsbeatgans,rombach2022highresolutionimagesynthesislatent}, driven by the development of diffusion-based models such as Seedance~\cite{seedance2025}, and HunyuanVideo~\cite{kong2024hunyuanvideo}.
Specifically, I2V generation transforms static images into temporally coherent videos, pushing generative models beyond single-image synthesis.
By drastically lowering the barrier of video creation, this technology facilitates diverse applications in entertainment~\cite{entertainment2019wen,entertainment2024BarLumiere}, education~\cite{AIforeducation2024}, and content generation~\cite{2018videogenerationcontentgeneration}.
Ultimately, the ability to produce high-quality videos from simple inputs democratizes production, empowering individual creators and reshaping creative industries.
Despite being powerful, I2V models may also introduce significant security risks, including the generation of deepfakes~\cite{pei2024deepfakegenerationdetectionbenchmark, westerlund2019emergence} and the dissemination of misinformation~\cite{NEWMAN2024101778misinformation}.

To mitigate these threats, prior research has proposed various image protection methods designed to prevent unauthorized content generation by I2V models~\cite{liang2023mist}.
Most existing methods adopt perturbation-based strategies, such as I2VGuard~\cite{i2vguard} and VGMShield~\cite{pang2025vgmshieldmitigatingmisusevideo}.
These methods build upon earlier studies that aim to protect images from unauthorized editing by Image-to-Image (I2I) diffusion models, including PhotoGuard~\cite{salman2023photoguard}, Mist~\cite{liang2023mist}, and EditShield~\cite{chen2024editshieldprotectingunauthorizedimage}.

Despite the emergence of these image protection methods, the field still lacks a common basis for evaluating them, and current assessments fall short in three respects. First, different methods are reported under different metrics and generation settings, so their protection effectiveness cannot be compared on common ground~\cite{ye2025evaluatingadversarialprotectionsdiffusion}.Second, many methods report protection only on the single target model they were optimized against, which leaves untested whether that protection generalizes to other, unseen I2V systems. Third, evaluations typically rely on a narrow, single-domain image set that does not cover the varied content types through which real misuse occurs.
To address these gaps, we introduce IPV-Bench (Image Protection against Video generation), the first systematic benchmark for image protection methods in I2V generation scenarios. IPV-Bench evaluates every method under one protocol along both fidelity and effectiveness, across I2V models that span distinct architectures as well as open-source and commercial closed-source systems, and over a data framework covering five misuse-relevant content domains, so that protection methods can be compared fairly and comprehensively.

We assess 5 protection methods (including both I2V-specific and I2I-based approaches) across 4 widely used I2V models featuring distinct structures, such as DiT and U-Net~\cite{peebles2023scalablediffusionmodelstransformers, ronneberger2015unetconvolutionalnetworksbiomedical}.
As illustrated in Figure~\ref{fig:mainfigure}, our benchmark offers a systematic assessment framework covering three dimensions: protection effectiveness, visual fidelity preservation, and robustness against preprocessing attacks.
Our evaluation delivers a consistently sobering picture.
\textit{Effectiveness}: under a unified $\epsilon{=}16/255$ budget, only 4 of 20 method--model pairs produce FVD above the per-model noise floor, and the sole broadly disruptive method (Mist) does so at the cost of severe image degradation (PSNR $27.1$), exposing a fundamental fidelity--effectiveness trade-off that no existing method resolves.
\textit{Content-dependence}: image content strongly governs the perceptual \emph{cost} of protection (LPIPS varies $43\%$ across domains through a spectral mechanism) but not its \emph{benefit}: no domain is systematically easier to protect, and the three effectiveness metrics rank domains inconsistently.
\textit{Transferability}: no perturbation clears the noise floor on more than two of the four generators; a budget-matched random perturbation matches or outperforms all learned defenses on two models, and no method affects the commercial Seedance API.
\textit{Robustness}: on the high-motion subset where protection should be most salient, 12 of 15 method--model pairs produce no measurable effect under clean conditions; of the three that do, Mist's protection on SVD collapses entirely under all three lightweight preprocessing transforms (H.264, JPEG, Gaussian noise), while only I2VGuard on Wan survives all attacks.
Together, these findings show that no current perturbation-based defense is simultaneously effective, architecture-agnostic, and robust to routine image operations.

To summarize, we make the following contributions:
\begin{itemize}
    \item \textbf{Benchmark.} We introduce IPV-Bench, the first systematic benchmark for evaluating image protection methods in I2V generation scenarios, together with IPV-500, the first prompt-paired image dataset purpose-built for this task. IPV-500 covers five misuse-relevant domains (faces, artworks, natural scenes, objects, animals) with motion-oriented video prompts generated and quality-scored by multiple VLMs, providing a controlled, reproducible evaluation foundation.

    \item \textbf{Key findings.} Our comprehensive evaluation reveals that no existing perturbation-based defense simultaneously preserves image fidelity, disrupts video generation, and survives lightweight preprocessing. Effective protection is rare, with only 4 of 20 method--model pairs exceeding the noise floor, and even those collapse under simple transforms such as JPEG compression or H.264 re-encoding. No method transfers reliably across I2V architectures, and a budget-matched random perturbation matches or outperforms learned defenses on multiple models. We further show that image content shapes only the perceptual cost of protection, not its benefit, so no domain offers an easier target.

    \item \textbf{Extensive empirical study.} To establish these results, we generate over 22,000 videos spanning five protection methods, four I2V models (including both open-source and commercial black-box APIs), and three robustness transforms. This scale makes IPV-Bench the most comprehensive evaluation of I2V protection to date.
\end{itemize}

\begin{figure*}[t]
\centering
\includegraphics[width=\textwidth]{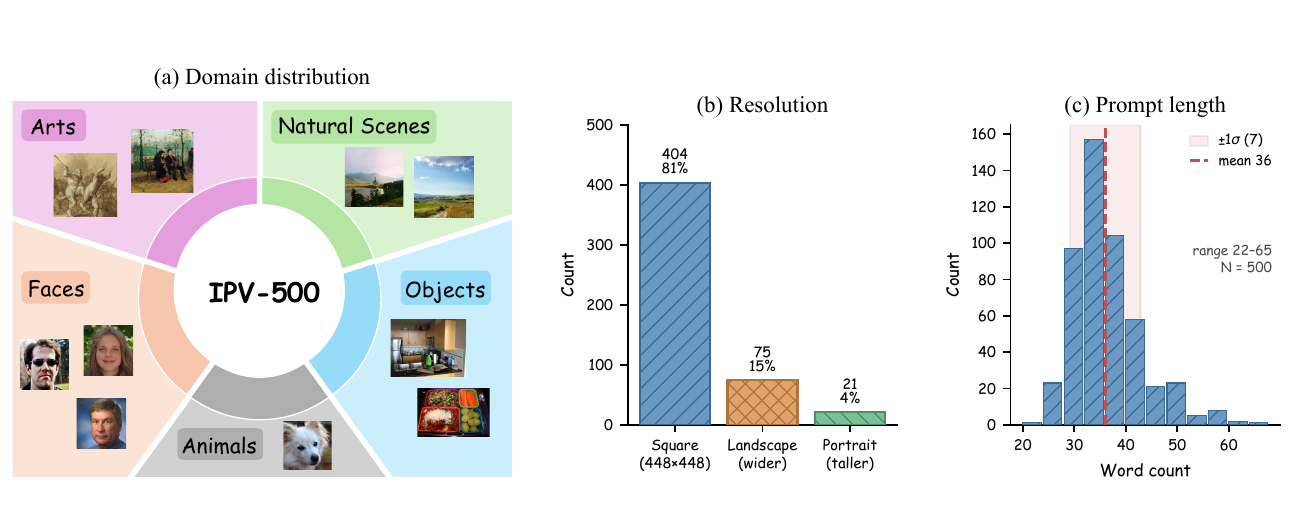}
\caption{\textbf{The IPV-500 dataset.} (a)~Domain distribution across five misuse-relevant categories (100 images each): faces, artworks, natural scenes, objects, and animals. (b)~Resolution distribution: 81\% of images are square (448$\times$448); the remainder are landscape or portrait, obtained by short-side resizing to preserve aspect ratio. (c)~Prompt word-count distribution for the 500 motion-oriented video prompts: each image is captioned by three VLMs (Qwen3-VL-4B, Qwen2.5-VL-7B, InternVL2-4B) and the best-scoring candidate is selected via CLIP-alignment, coherence, and inter-prompt diversity scoring. Mean 36 words, range 22--65.}
\label{fig:dataset}
\end{figure*}

\section{Related Work}

\mypara{Perturbation-based Image Protection}
Imperceptible adversarial perturbations have become the dominant paradigm for protecting images against diffusion-based misuse~\cite{choi2025diffusionguardrobustdefensemalicious, ye2025evaluatingadversarialprotectionsdiffusion, chowdhury2025vidfreezeprotectingimagesmalicious}, differing mainly in where they intervene in the generative pipeline. For Image-to-Image (I2I) editing and personalization, PhotoGuard~\cite{salman2023photoguard} perturbs the diffusion-encoder latent, Mist~\cite{liang2023mist} adds a semantic denoising loss, EditShield~\cite{chen2024editshieldprotectingunauthorizedimage} optimizes over transformations, and Glaze~\cite{shan2025glazeprotectingartistsstyle} and Anti-DreamBooth~\cite{vanle2023antidreambooth} disrupt style mimicry and personalization. As misuse extended to video, temporally-aware defenses followed: I2VGuard~\cite{i2vguard} targets temporal-attention modules, and VGMShield~\cite{pang2025vgmshieldmitigatingmisusevideo} attacks the conditioning embedder. These methods are hard to compare on equal terms: each is tuned to its own budget, validated on a single generator or task, and reported with a metric of its own choosing. IPV-Bench targets this fragmentation, evaluating representative I2I- and I2V-oriented defenses under a shared budget, across multiple generators, and along common axes.

\mypara{Robustness of Protective Perturbations}
A protective perturbation is useful only if it survives the processing an image undergoes before reaching a generator. Prior work repeatedly shows such perturbations are brittle: simple rule-based transformations substantially weaken them~\cite{ye2025evaluatingadversarialprotectionsdiffusion}, echoing watermark and adversarial-robustness benchmarks that protection claims must hold under distribution shift~\cite{an2024waves, croce2021robustbench}. The threats fall into two families: computationally intensive diffusion-based purification (e.g., DiffPure~\cite{nie2022diffpure}), and lightweight rule-based transformations such as compression and additive noise applied at scale. Existing studies probe robustness only narrowly, against a single transformation or none, leaving open how defenses withstand the everyday low-cost operations images encounter on sharing platforms. IPV-Bench integrates robustness as a core axis, benchmarking these practical transformations while treating heavier purification as a complementary higher-cost threat.

\mypara{Image and Video Quality Evaluation}
Image protection is two-sided: a perturbation must stay faithful to the source image while degrading the video generated from it, requiring measurement on both fronts. For the protected image, evaluations rely on PSNR and SSIM~\cite{wang2004image}, LPIPS~\cite{zhang2018unreasonable}, and CLIP~\cite{radford2021learning}. For generation, benchmarks decompose video quality into fine-grained axes, e.g., VBench~\cite{huang2023vbench, huang2025vbench++} and EvalCrafter~\cite{liu2024evalcrafter}, while HEIM~\cite{lee2023holistic} argues for holistic assessment and FVD~\cite{unterthiner2019accurategenerativemodelsvideofvd} captures distributional shift, spanning backbones from U-Net-based SVD~\cite{blattmann2023SVD} to DiT models such as Wan2.2~\cite{wan2025wanopenadvancedlargescale} and SkyReel~\cite{chen2025skyreelsv2infinitelengthfilmgenerative} and commercial APIs. These benchmarks, however, evaluate \emph{the generator}, not the protection; existing methods in turn report effectiveness in isolation, each on an ad hoc metric and generator, so the field lacks a unified answer to how well defenses work. IPV-Bench fills this gap with a multi-dimensional suite spanning image fidelity (PSNR, SSIM, LPIPS, CLIP) and video disruption (FVD, VBench, prompt-video alignment). By shifting the object of study from the generator to the \emph{perturbation itself} and scoring both axes jointly, IPV-Bench offers the first unified comparison of image-protection methods for I2V generation.

\section{IPV-Bench}
\label{sec:benchmark}

We describe IPV-Bench in three parts: a threat model that formalizes the image-protection task, the IPV-500 dataset that supplies realistic and prompt-paired inputs, and an evaluation protocol that scores each method jointly on image fidelity and video disruption. Figure~\ref{fig:mainfigure} gives an overview of the full pipeline.

\subsection{Threat Model}

We cast image protection as an interaction between a \emph{defender}, who owns a source image $x$, and an \emph{adversary}, who seeks to generate video from it with an image-to-video model $G$. Before releasing the image, the defender adds an imperceptible perturbation $\delta$ to obtain a protected image
\begin{equation}
    x' = x + \delta, \qquad \lVert \delta \rVert_\infty \le \epsilon ,
\end{equation}
which should remain visually faithful to $x$ while degrading the video that $G$ produces from it. The defender therefore seeks a perturbation that maximizes the discrepancy between the videos generated from the protected and clean images,
\begin{equation}
    \max_{\delta}\; \mathcal{D} \!\big(G(x'),\, G(x)\big) \quad \text{s.t.} \quad \lVert \delta \rVert_\infty \le \epsilon ,
\end{equation}
where $\mathcal{D}$ denotes a measure of video-level discrepancy.

The adversary observes only the released image $x'$ and is \emph{black-box} with respect to the protection algorithm: it knows neither $\delta$ nor the method that produced it. To neutralize the perturbation, the adversary may apply a low-cost preprocessing transform $T$ before generation, producing $G(T(x'))$. Protection is thus successful only when the generated video remains degraded, both directly and after such preprocessing, while $x'$ stays perceptually close to $x$.

\subsection{The IPV-500 Dataset}

Image protection matters across visual domains that differ sharply in structure and misuse risk, and real images arrive in heterogeneous shapes rather than lab-cropped squares. IPV-500 is built around both observations. It contains 500 images evenly spread over five misuse-relevant domains ($100$ each): \textbf{faces} (FFHQ~\cite{karras2019stylebasedgeneratorarchitecturegenerative}) for impersonation risk, \textbf{artworks} (\cite{wikiart}) for style theft, \textbf{animals} (AFHQ-v2) as non-human subjects, \textbf{natural scenes} (LHQ) for scene-level dynamics, and \textbf{objects} (MSCOCO~\cite{lin2015microsoftcococommonobjects}) for object-centric generation. Faces, artworks, and animals are standardized to $448{\times}448$, whereas scenes and objects retain their native aspect ratio (short side $448$) to mirror real social-media uploads.

\mypara{Construction}
Rather than maximize scale, IPV-500 prioritizes quality and coverage. Candidate images are oversampled per domain and filtered by a vision-language quality gate, and each retained image is paired with a motion-oriented video prompt produced by multiple vision-language models and screened by automatic quality checks with human adjudication of uncertain cases. For robustness experiments, we further select a $150$-image high-motion subset by ranking images on optical-flow dynamic degree and inter-frame CLIP drift, since perturbation effects are most salient where genuine motion is expected. Figure~\ref{fig:dataset} summarizes the resulting distribution.

\subsection{Evaluation Protocol}

\mypara{Subjects under test}
IPV-Bench evaluates five representative protection methods (the I2I-origin PhotoGuard~\cite{salman2023photoguard}, Mist~\cite{liang2023mist}, and EditShield~\cite{chen2024editshieldprotectingunauthorizedimage}, and the I2V-specific I2VGuard~\cite{i2vguard} and VGMShield~\cite{pang2025vgmshieldmitigatingmisusevideo}), together with a RandomNoise reference that anchors how much of any observed disruption is attributable to arbitrary perturbation. Each is applied across four I2V systems that span both architectures and access modes: the U-Net--based SVD-XT~\cite{blattmann2023SVD}; the Diffusion-Transformer models Wan2.2-TI2V-5B~\cite{wan2025wanopenadvancedlargescale} and SkyReels-V2-I2V~\cite{chen2025skyreelsv2infinitelengthfilmgenerative}; and the commercial black-box API Seedance~\cite{seedance2025}. Every method is configured to a common budget of $\epsilon = 16/255$, so that measured differences reflect how a method allocates its perturbation rather than its magnitude.

\mypara{Image fidelity}
We first measure how perceptible a protective perturbation is, i.e., how much usability the defender sacrifices. We report PSNR and SSIM~\cite{wang2004image} for structural fidelity, LPIPS~\cite{zhang2018unreasonable} for perceptual similarity, and CLIP image similarity~\cite{radford2021learning} for semantic preservation.

\mypara{Protection effectiveness}
We then measure how much a protected image disrupts generation, using three metrics that capture complementary failure modes. Fr\'{e}chet Video Distance (FVD)~\cite{unterthiner2019accurategenerativemodelsvideofvd}, our primary scalar, measures the distributional shift between videos generated from clean and protected images; a CLIP text--video degradation rate measures loss of prompt--video semantic alignment; and VBench~\cite{huang2023vbench} degradation profiles video quality along six complementary dimensions (subject consistency, background consistency, motion smoothness, temporal flickering, aesthetic quality, and imaging quality), spanning both frame-level appearance and temporal behavior. Effectiveness on each metric is expressed as a degradation rate $\mathrm{DR} = (S_{\text{orig}} - S_{\text{prot}}) / S_{\text{orig}}$, so that positive values denote degradation (successful protection) and negative values denote a protected video that scores \emph{better} than the clean one. Because the six VBench dimensions occupy very different value ranges, where a raw average is dominated by the few high-variance ones, we scale-normalize each dimension by its across-grid standard deviation before averaging, which equalizes the dimensions' contributions while preserving the sign and zero point of the DR. We cross-check conclusions across all three metrics rather than trusting any one.

\mypara{Robustness}
Finally, we test whether protection survives the preprocessing an adversary may apply before generation. Instantiating the transform $T$ from our threat model, we subject each protected image to three image preprocessing attacks (JPEG compression, additive Gaussian noise, and an H.264 codec round-trip that approximates video-platform artifacts), and re-measure effectiveness to gauge each method's robustness to them. To prevent conflating protection with videos that are inherently static, we run these experiments on the high-motion subset introduced above.

\subsection{Evaluation Noise Floor}
\label{sec:floor}
FVD is a finite-sample estimator: with a few hundred videos and $1024$-dimensional I3D features, even two video sets drawn from the \emph{same} distribution yield a nonzero FVD purely from sampling variance. Reading a raw FVD as ``protection strength'' without accounting for this variance can therefore mistake estimator noise for a real effect. To calibrate what magnitude of FVD is meaningful, we measure a per-model \emph{noise floor}: we bootstrap two independent resamples of the \emph{unprotected} videos and compute their FVD, repeating this to obtain the distribution of FVD expected under \emph{zero} true distributional shift. We report the $95$th percentile of this distribution as the floor. As a sanity check, the FVD of a video set against itself is exactly $0$, confirming the estimator is unbiased at identity. A protected set is credited with a genuine distributional shift only when its FVD exceeds this floor; values within the floor are statistically indistinguishable from applying no protection at all. This noise floor complements the RandomNoise baseline of \S\ref{sec:benchmark}: the baseline asks whether a method outperforms an arbitrary perturbation of equal budget, while the floor asks whether an observed FVD is distinguishable from measurement noise at all.

\section{Experiments}

\subsection{Experimental Setup}
We evaluate the five protection methods and the RandomNoise baseline of \S\ref{sec:benchmark} on IPV-500, generating videos with all four I2V systems and scoring them with the fidelity and effectiveness metrics of \S\ref{sec:floor}. Throughout, we interpret every FVD against two reference lines: the RandomNoise baseline (is a method better than an arbitrary perturbation of equal budget?) and the per-model noise floor (is an FVD distinguishable from estimator noise at all?). All numbers use the full 500-image set except the robustness study, which uses the 150-image high-motion subset. We organize our analysis around four findings.

\begin{table}[t]
\centering
\small
\setlength{\tabcolsep}{5pt}
\renewcommand{\arraystretch}{1.1}
\caption{\textbf{Image fidelity of protected images} (model-independent). Arrows mark the imperceptible direction.}
\label{tab:fid}
\begin{tabular}{lcccc}
\toprule
Method & PSNR\,$\uparrow$ & SSIM\,$\uparrow$ & LPIPS\,$\downarrow$ & CLIP-I\,$\uparrow$ \\
\midrule
PhotoGuard & 31.9 & 0.819 & 0.200 & 0.950 \\
Mist & 27.1 & 0.713 & 0.358 & 0.897 \\
I2VGuard & \textbf{35.1} & \textbf{0.929} & \textbf{0.093} & \textbf{0.967} \\
VGMShield & 31.8 & 0.812 & 0.243 & 0.929 \\
EditShield & 34.3 & 0.870 & 0.180 & 0.961 \\
\midrule
RandomNoise & 28.9 & 0.688 & 0.196 & 0.963 \\
\bottomrule
\end{tabular}
\end{table}

\begin{table*}[t]
\centering
\small
\setlength{\tabcolsep}{5pt}
\renewcommand{\arraystretch}{1.1}
\caption{\textbf{Protection effectiveness per I2V model} under a unified $L_\infty=16/255$ budget. All three metrics are degradation rates where \emph{higher $=$ stronger protection} ($\uparrow$): FVD; scale-normalized VBench degradation VB (std.\ units); CLIP-T degradation CT (\%). A value ${\le}0$ means no degradation. \textbf{Bold} marks entries above the corresponding reference: FVD above the per-model noise floor (\emph{floor} row, \S\ref{sec:floor}); VB and CT above the RandomNoise baseline. Seedance is a commercial black-box API.}
\label{tab:eff}
\begin{tabular}{l ccc ccc ccc ccc}
\toprule
& \multicolumn{3}{c}{\textbf{Wan}} & \multicolumn{3}{c}{\textbf{SkyReel}} & \multicolumn{3}{c}{\textbf{SVD}} & \multicolumn{3}{c}{\textbf{Seedance}} \\
\cmidrule(lr){2-4}\cmidrule(lr){5-7}\cmidrule(lr){8-10}\cmidrule(lr){11-13}
Method & FVD\,$\uparrow$ & VB\,$\uparrow$ & CT\,$\uparrow$ & FVD\,$\uparrow$ & VB\,$\uparrow$ & CT\,$\uparrow$ & FVD\,$\uparrow$ & VB\,$\uparrow$ & CT\,$\uparrow$ & FVD\,$\uparrow$ & VB\,$\uparrow$ & CT\,$\uparrow$ \\
\midrule
PhotoGuard & 3.8 & +0.67 & \textbf{$-$0.4} & 4.9 & +0.49 & $-$0.2 & 4.6 & \textbf{+0.94} & +0.8 & 5.2 & \textbf{+0.56} & $-$0.5 \\
Mist & \textbf{8.5} & \textbf{+3.40} & \textbf{+0.4} & 5.9 & +1.04 & $-$0.1 & \textbf{8.5} & \textbf{+2.14} & \textbf{+4.7} & 5.7 & \textbf{+1.26} & $-$1.0 \\
I2VGuard & \textbf{7.3} & \textbf{+2.61} & \textbf{+3.3} & 4.4 & +0.21 & $-$0.1 & 4.3 & \textbf{+0.57} & \textbf{+1.1} & 5.2 & +0.31 & $-$0.5 \\
VGMShield & 4.6 & +0.71 & \textbf{$-$0.3} & 5.2 & +0.64 & +0.2 & \textbf{5.6} & \textbf{+1.08} & \textbf{+2.1} & 5.3 & \textbf{+0.80} & $-$0.8 \\
EditShield & 3.2 & +0.09 & \textbf{+0.1} & 4.6 & +0.30 & $-$0.1 & 4.4 & \textbf{+0.52} & +0.2 & 5.3 & \textbf{+0.50} & $-$0.7 \\
\midrule
RandomNoise & \textbf{6.0} & +1.82 & -0.6 & \textbf{7.4} & +1.15 & +0.4 & 4.0 & +0.07 & +0.9 & 5.3 & +0.50 & -0.4 \\
\textit{noise floor} & \textit{5.8} & -- & -- & \textit{6.2} & -- & -- & \textit{5.2} & -- & -- & \textit{5.9} & -- & -- \\
\bottomrule
\end{tabular}
\end{table*}

\begin{figure}[t]
\centering
\includegraphics[width=0.92\columnwidth]{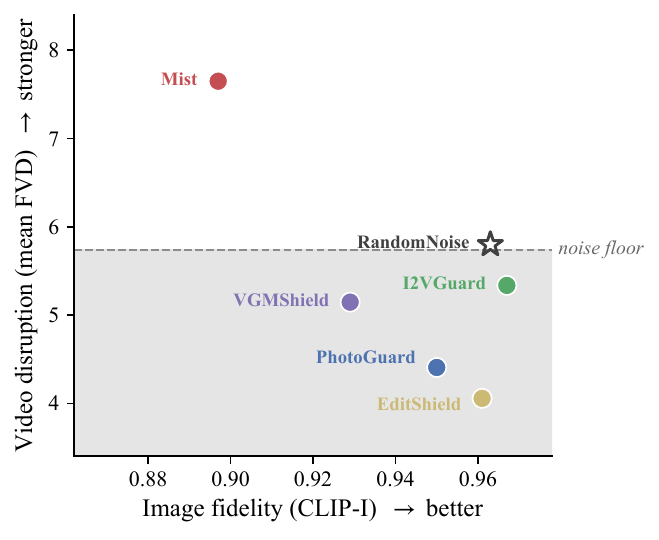}
\caption{\textbf{The fidelity--effectiveness trade-off.} Methods that preserve the image barely disrupt generation, and the only disruptive method sacrifices fidelity.}
\label{fig:pareto}
\end{figure}

\subsection{No Method Balances Fidelity and Protection}
\textbf{Finding 1: under a unified budget, image protection and video disruption trade off against each other, and most methods fail to disrupt generation beyond noise.}
Tables~\ref{tab:fid} and~\ref{tab:eff} together expose a stark trade-off. The methods that best preserve the image are the weakest protectors: I2VGuard attains the highest fidelity (PSNR $35.1$, LPIPS $0.09$) and EditShield is close behind, yet on most generators their FVD sits \emph{below} both the RandomNoise baseline and the noise floor, i.e., their protected videos are statistically indistinguishable from unprotected ones. Conversely, the only method with broadly non-trivial FVD, Mist, does so with the worst fidelity by a wide margin (PSNR $27.1$, LPIPS $0.36$), degrading the released image far more than any other method. No method occupies the desirable corner of high fidelity \emph{and} strong disruption; Figure~\ref{fig:pareto} shows this frontier directly.

Beyond the trade-off, the disruption that methods do achieve is weak in absolute terms. Across Wan, SkyReel, SVD, and the commercial Seedance, only four of the twenty method--model combinations clear their model's noise floor (Mist on Wan and SVD, I2VGuard on Wan, and VGMShield marginally on SVD), and even these exceed it by only $1.1$--$1.6\times$; the remaining sixteen are statistically indistinguishable from applying no protection at all. In other words, under a fair budget and after accounting for estimator noise, the dominant outcome is \emph{no measurable protection}. We next ask whether image content shifts this cost--benefit balance.

\begin{figure*}[t]
\centering
\includegraphics[width=0.84\textwidth]{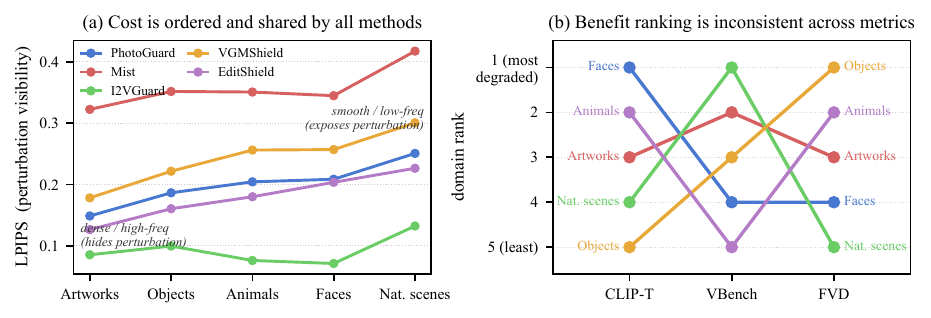}
\caption{\textbf{Image content shapes the cost of protection, not its benefit.} (a) Perturbation visibility (LPIPS, averaged over methods) rises from dense, high-frequency domains (artworks) to smooth, low-frequency ones (natural scenes), and \emph{every} method follows the same ordering. (b) The three effectiveness metrics rank the five domains inconsistently (crossing lines): CLIP-T ranks faces first, VBench natural scenes, FVD objects. The per-domain FVD ordering is reproduced almost exactly by a null-protection method, so it tracks intrinsic domain diversity rather than protection. Effectiveness magnitudes are near-zero throughout (CLIP-T degradation $\leq 1\%$ in every domain).}
\label{fig:domain}
\end{figure*}

\subsection{Content Shapes the Cost of Protection, Not Its Benefit}
\textbf{Finding 2: image content strongly governs how \emph{visible} a protective perturbation is, but not how \emph{effective} it is: no domain is systematically easier to protect.}
Holding the budget and method fixed and varying only the image domain (the five IPV-500 categories, $100$ images each) separates the two sides of the Finding-1 trade-off. On the cost side, visibility is strongly content-dependent: averaged over methods, LPIPS swings $43\%$ across domains, from $0.17$ on artworks to $0.27$ on natural scenes, and \emph{every} method induces the same ordering (Figure~\ref{fig:domain}a). The mechanism is spectral: dense, high-frequency domains (artworks, cluttered object scenes) mask a fixed-budget perturbation, whereas smooth, low-frequency domains (open landscapes, faces) expose it, mirroring human sensitivity to perturbations in flat regions. Content-awareness thus offers a real lever on imperceptibility.

It offers none on protection. The CLIP-T degradation reaches at most $1\%$ in any domain (from $-0.3\%$ on objects to $+1.0\%$ on faces), and the three effectiveness metrics disagree entirely on which domain is best protected: CLIP-T ranks faces first, VBench ranks natural scenes first, and FVD ranks objects first (Figure~\ref{fig:domain}b). This disorder reflects the absence of signal rather than measurement noise: the per-domain FVD ordering, in particular, is reproduced almost exactly by a \emph{null}-protection method (Pearson $0.90$), so it tracks the intrinsic visual diversity of each domain, not any protective effect. No domain yields a systematically more disruptable target. Content therefore sets only the price a defender pays in image quality, not the protection obtained, which removes the option of escaping the trade-off by choosing a favorable domain.

\subsection{Protection Does Not Generalize Across Generators}
\begin{figure}[t]
\centering
\includegraphics[width=0.82\columnwidth]{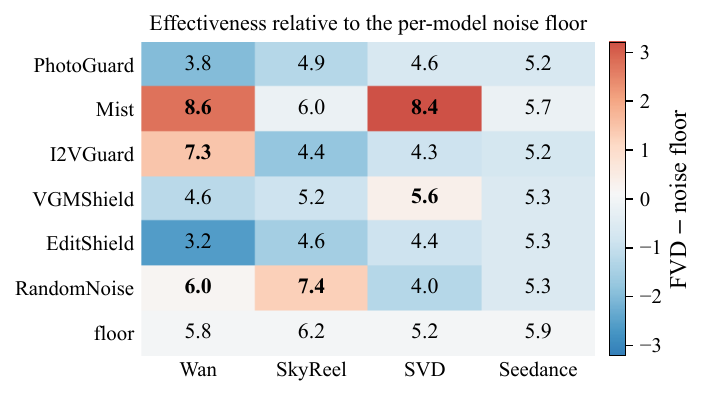}
\caption{\textbf{Protection does not generalize across generators.} Effectiveness above the noise floor is a scattered, model-specific accident, uncorrelated with each method's design target.}
\label{fig:transfer}
\end{figure}

\textbf{Finding 3: no perturbation reliably transfers across I2V models: methods fail even on their own target model, and a random perturbation generalizes as well as any learned one.}
Reading Table~\ref{tab:eff} across its columns exposes a further failure. Of the method--model combinations, only isolated cells clear their model's noise floor, and \emph{no method clears it on more than two of the models}. The pattern is uncorrelated with algorithmic design: I2VGuard and VGMShield are optimized directly against SVD, yet I2VGuard does not clear the floor on SVD at all ($4.3$ vs.\ $5.25$) while it does on Wan, a model it never targeted; the strongest method, Mist, is an I2I method that happens to clear the floor on Wan and SVD but not SkyReel. Most tellingly, the RandomNoise baseline clears the floor on two models (Wan and SkyReel) and is the \emph{only} perturbation to affect SkyReel at all, so a budget-matched random perturbation transfers at least as broadly as any learned defense. No method produces a noise-distinguishable shift on the commercial Seedance API. Because a protective perturbation is optimized against one surrogate encoder, it neither reliably disrupts its own target generator nor transfers to unseen ones (Figure~\ref{fig:transfer}); reliable protection would require perturbations that generalize across architectures rather than overfitting a single surrogate.

\begin{figure}[t]
\centering
\includegraphics[width=\columnwidth]{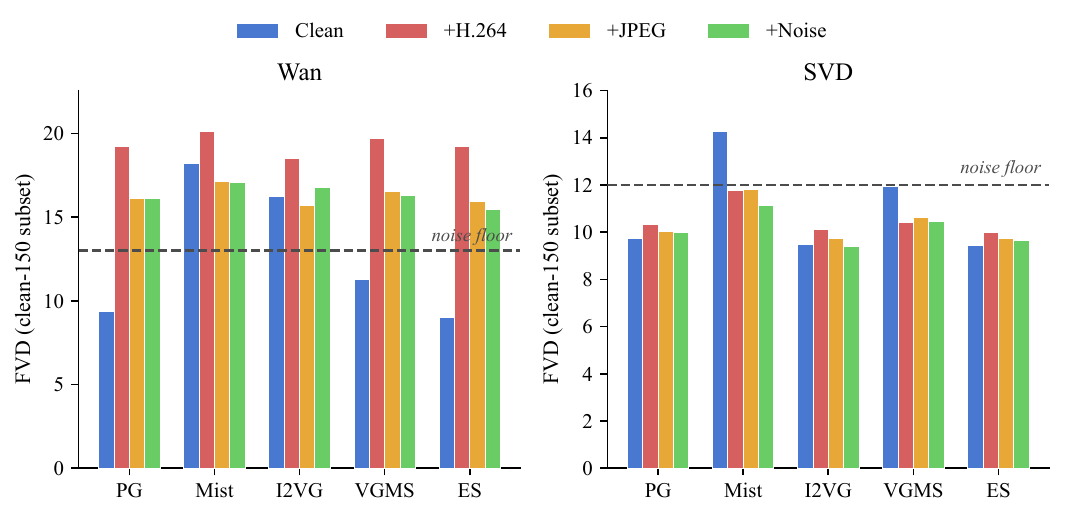}
\caption{\textbf{Effective protection is rare and brittle (high-motion subset, $N{=}150$).} On Wan and SVD, the only two models where any method clears the noise floor (dashed line), Mist is the sole method with measurable clean-setting effect, and its protection on SVD collapses below the floor under all three preprocessing variants. I2VGuard on Wan is more resilient but still the exception rather than the rule.}
\label{fig:robust}
\end{figure}

\subsection{Effective Protection Is Rare and Brittle}
\textbf{Finding 4: on a high-motion subset where protection should be most salient, most methods produce no measurable effect; the few that do are broken by simple preprocessing.}

We evaluate robustness on the 150-image high-motion subset, which was specifically selected to maximize the signal-to-noise ratio for protection experiments: high-motion videos yield larger FVD shifts and are thus where adversarial perturbations should appear most effective.
Despite this favorable setting, the results expose a two-layer failure.

\mypara{Layer 1 -- most methods fail even without attack}
On the high-motion subset under clean (no-attack) conditions, only three method--model pairs produce FVD above the noise floor: Mist on Wan ($18.2$ vs.\ floor $13.0$), Mist on SVD ($14.3$ vs.\ floor $12.0$), and I2VGuard on Wan ($16.2$ vs.\ floor $13.0$).
The remaining 12 of 15 pairs fall below their respective floors, indistinguishable from no protection at all.
We attribute part of this to our more demanding generation setting: while prior work validates protection on SVD with 25 frames (${\approx}1$s), our benchmark uses up to 49 frames (SkyReel, ${\approx}6$s), and tests across DiT-based architectures (Wan, SkyReel) that no existing method was designed against.

\mypara{Layer 2 -- the few effective cases are fragile}
Among the three method--model pairs that do clear the noise floor, all three are undermined by at least one preprocessing transform (Figure~\ref{fig:robust}).
Mist's protection on SVD collapses under every attack variant: H.264 ($11.8 < 12.0$), JPEG ($11.8 < 12.0$), and Gaussian noise ($11.1 < 12.0$), pushing FVD below the floor in all three cases.
Mist on Wan is partially reduced by JPEG and noise (FVD drops from $18.2$ to ${\approx}17.1$), though it remains above the floor.
I2VGuard on Wan is more resilient, with FVD staying above the floor across all three attacks ($15.7$--$18.5$ vs.\ floor $13.0$).

These results reveal that current protections are optimized as high-spatial-frequency signals tuned to a single surrogate encoder, precisely what JPEG compression and Gaussian noise discard first. An adversary requires no knowledge of the specific defense to neutralize it. Future work should pursue frequency-aware and multi-surrogate perturbation strategies to address this brittleness.

\section{Conclusion}
We introduced IPV-Bench, the first systematic benchmark for evaluating image protection methods against I2V generation, comprising IPV-500 and a unified evaluation protocol spanning five methods, four I2V models, and three dimensions: protection effectiveness, visual fidelity, and robustness against preprocessing.
Our results consistently reveal a gap between the promise and the practice of current defenses: no method simultaneously preserves image fidelity and disrupts generation; protection effects neither transfer across architectures nor survive simple preprocessing.
Together, these findings suggest that the field has not yet produced a perturbation-based defense that is simultaneously effective, architecture-agnostic, and robust to routine image operations.
We hope IPV-Bench provides a rigorous foundation for the next generation of protection methods and lowers the barrier to systematic evaluation as new generators and defenses emerge.

% ---------------------------------------------------------------
% ---------------------------------------------------------------
\bibliographystyle{plain}
\bibliography{main}

\clearpage
\appendix
\onecolumn

\section{IPV-500 Dataset Construction}
\label{app:dataset}

This appendix documents the full construction of IPV-500 so that the dataset can be reproduced or extended. We describe (i) the source corpora, (ii) the end-to-end curation pipeline with all prompts and hyperparameters, and (iii) representative image--prompt examples.

\subsection{Data Sources}
IPV-500 draws 100 images from each of five misuse-relevant domains, sampled from established public corpora. Domains are balanced by design (100 images each, 500 total) so that no single content type dominates the aggregate statistics. Table~\ref{tab:sources} lists the source corpus and native characteristics of each domain.

\begin{table}[h]
\centering
\small
\setlength{\tabcolsep}{6pt}
\renewcommand{\arraystretch}{1.2}
\caption{Source corpora for the five IPV-500 domains. Each domain contributes exactly 100 images.}
\label{tab:sources}
\begin{tabular}{lllc}
\toprule
Domain & Source corpus & Content & Aspect handling \\
\midrule
Faces          & FFHQ~\cite{karras2019stylegan}          & Human portraits          & center-crop square \\
Artworks       & \cite{wikiart}                          & Paintings, drawings      & center-crop square \\
Natural scenes & LHQ~\cite{skorokhodov2021aligning}      & Landscapes               & short-side resize \\
Objects        & MS-COCO~\cite{lin2015microsoftcococommonobjects} & Everyday object scenes   & short-side resize \\
Animals        & AFHQ-v2~\cite{choi2020stargan}          & Animal faces             & center-crop square \\
\bottomrule
\end{tabular}
\end{table}

All images are standardized to a short edge of $448$\,px. Domains with an inherently square framing (faces, artworks, animals) are center-cropped to $448{\times}448$; domains where aspect ratio carries semantic content (natural scenes, objects) are resized by the short side to preserve their native layout, yielding the resolution distribution reported in Figure~\ref{fig:dataset}(b).

\subsection{Curation Pipeline}
\label{app:pipeline}
Each retained image is paired with a \emph{motion-oriented} video prompt. The pipeline has four stages: image quality screening, multi-VLM prompt generation, automatic quality control (QC), and human adjudication of borderline cases.

\paragraph{Stage 1: Image quality screening.}
Before prompt generation, each raw candidate image is audited by a Qwen2.5-VL-7B-Instruct classifier to ensure a clearly visible, adequately framed main subject free of severe artifacts. The auditor uses the following prompt:

\begin{promptbox}[VLM image quality auditor prompt (Qwen2.5-VL-7B-Instruct)]
\small\rmfamily
You are an image quality auditor. Check if this image has:\\
\hspace*{1em}(a) a clearly visible main subject (person/animal/object/scene),\\
\hspace*{1em}(b) no severe cropping of the main subject ($>$50\% missing),\\
\hspace*{1em}(c) no severe artifacts (heavy blur, large watermarks).\\
Answer \textbf{PASS} or \textbf{FAIL}. If FAIL, give the reason in $\leq$10 words.
\end{promptbox}

Images receiving a FAIL verdict are discarded and replaced by the next candidate from the source corpus. Each subset is oversampled at $1.5\times$ the target (i.e.\ up to 150 candidates scanned per 100-image domain, fixed seed 42) to absorb rejections without manual intervention. Rejection reasons observed in practice included severe blur, absence of a discernible main subject, heavy cropping, and decorative or abstract content unsuitable for motion-oriented prompting.

\paragraph{Stage 2: Multi-VLM prompt generation.}
For every image we independently query three open vision-language models: Qwen3-VL-4B-Instruct, Qwen2.5-VL-7B-Instruct, and InternVL2-4B. Each model receives the image and the adversary-minded instruction shown below. The instruction deliberately models a misuse-minded user who fixates on the salient subject and animates it, so that the prompts represent the strongest realistic I2V-misuse scenario rather than benign captions. Generation is deterministic (greedy decoding, \texttt{do\_sample=False}, \texttt{max\_new\_tokens=180}); each model returns a list of salient entities and a single-line video prompt.

\begin{promptbox}[Adversary-minded image-to-video instruction (given to each VLM)]
\small\rmfamily
Turn this image into a striking video. First identify 1--3 salient entities, then animate them with vivid motion. If entities are subtle (portraits, landscapes), add environment dynamics like weather, lighting, or atmosphere. Format: one line, $\sim$80 words, no quotes.\\[2pt]
\textit{Reply as:} \quad Entities: \texttt{<e1>, <e2>} $\mid$ Prompt: \texttt{<video prompt>}\\[4pt]
\textit{Examples:}\\
Entities: tabby cat $\mid$ Prompt: The tabby cat's ears flick back, then it lunges forward, paw swiping at the lens as its mouth opens in a hiss before spinning and darting off-frame.\\
Entities: young woman $\mid$ Prompt: The woman's eyes widen as a slow smile spreads across her face, loose strands of hair lifting in a sudden breeze while soft golden light sweeps across the scene.\\
Entities: misty forest, distant peaks $\mid$ Prompt: Thick fog rolls through the forest in waves, shrouding the trees as rain begins to fall, droplets streaking down and blurring the distant peaks.
\end{promptbox}

\paragraph{Stage 3: Automatic quality control.}
The three candidate prompts for each image are scored on three complementary axes, and the best-scoring candidate that passes all thresholds is selected. Prompts failing any threshold are flagged for human review (Stage 3). The QC criteria are summarized below.

\begin{promptbox}[Automatic QC criteria for candidate prompts]
\small\rmfamily
Each candidate prompt is scored on three axes; a prompt passing \emph{all} thresholds becomes a \textsc{candidate}, otherwise it is \textsc{flagged} for human review.
\begin{itemize}\itemsep1pt \topsep1pt
  \item \textbf{Alignment.} CLIP image--text cosine similarity between the image and the prompt. Threshold $\geq 0.24$.
  \item \textbf{Coherence.} A readability heuristic on length and lexical structure, scaled to $[0,100]$. Threshold $\geq 91$.
  \item \textbf{Diversity.} Mean cosine distance of the prompt embedding to the other two candidates, discouraging near-duplicate phrasings. Threshold $\geq 0.04$.
\end{itemize}
Alignment uses the CLIP ViT-B/32 image and text encoders. Grounding via per-noun matching was intentionally omitted, since adversarial prompts emphasize dynamics and actions over static object lists, which makes per-noun overlap counterproductive.
\end{promptbox}

\paragraph{Stage 4: Human adjudication.}
Prompts flagged by QC are manually reviewed and either corrected or replaced by the highest-scoring alternative candidate. This step removes hallucinated entities, degenerate repetitions, and prompts that describe the wrong subject. After adjudication every image in IPV-500 carries exactly one validated video prompt; the resulting prompt-length distribution (mean $36$ words, range $22$--$65$) is shown in Figure~\ref{fig:dataset}(c).

\subsection{Image--Prompt Examples}
\begin{figure}[h]
\centering
\begin{promptbox}[Faces \texttt{(faces\_000)}]
\begin{minipage}[c]{0.18\textwidth}
  \includegraphics[width=\linewidth]{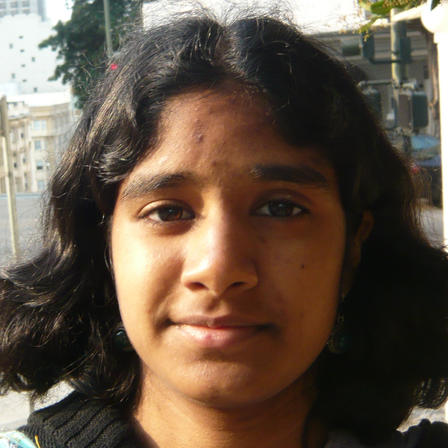}
\end{minipage}\hfill
\begin{minipage}[c]{0.79\textwidth}
  \footnotesize \textbf{Entities:} young woman, earrings, background cityscape.\\
  \textbf{Prompt:} \gentext{The young woman's earrings glisten as she turns her head, her hair swaying gently in a breeze, with the cityscape blurred in the background, creating a dynamic contrast between her and the urban environment.}
\end{minipage}
\end{promptbox}
\begin{promptbox}[Animals \texttt{(animals\_000)}]
\begin{minipage}[c]{0.18\textwidth}
  \includegraphics[width=\linewidth]{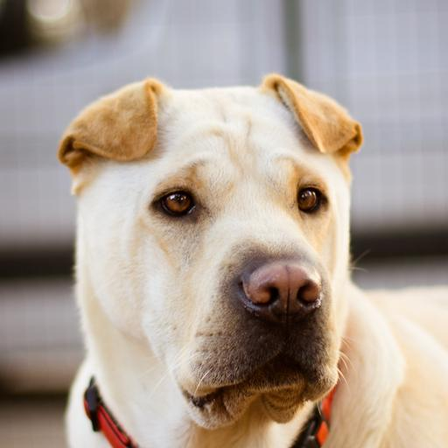}
\end{minipage}\hfill
\begin{minipage}[c]{0.79\textwidth}
  \footnotesize \textbf{Entities:} Labrador Retriever, dog collar, outdoor setting.\\
  \textbf{Prompt:} \gentext{The Labrador Retriever's head turns slightly to the side, its ears perked up, as it gazes intently at the camera. The dog's fur glistens under the soft, natural lighting, and the background subtly shifts to show a blurred outdoor setting with hints of greenery and a distant fence.}
\end{minipage}
\end{promptbox}
\begin{promptbox}[Natural scenes \texttt{(natural\_scenes\_000)}]
\begin{minipage}[c]{0.18\textwidth}
  \includegraphics[width=\linewidth]{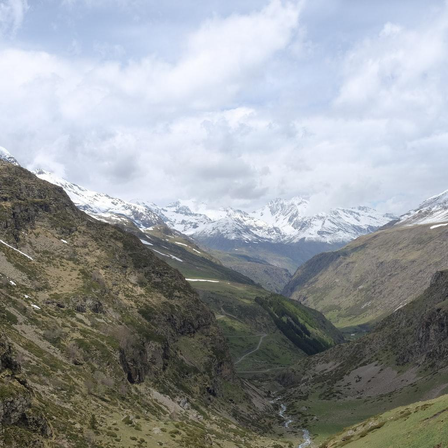}
\end{minipage}\hfill
\begin{minipage}[c]{0.79\textwidth}
  \footnotesize \textbf{Entities:} snow-capped mountains, winding valley path.\\
  \textbf{Prompt:} \gentext{Snow-dusted peaks loom above as a narrow valley path snakes through the terrain, its surface glistening with dew under shifting clouds, while wind whispers through the slopes, stirring grass and stone in slow, sweeping motion.}
\end{minipage}
\end{promptbox}
\caption{Representative IPV-500 image--prompt pairs (image left, model-generated prompt right). Prompts emphasize motion and environment dynamics, modeling a misuse-minded I2V user.}
\label{fig:app_examples}
\end{figure}

Figure~\ref{fig:app_examples} shows three representative image--prompt pairs, one per domain type (subject-centric, animal, scene-centric), illustrating the motion-oriented, adversary-style prompts produced by the pipeline.

\section{Generation and Protection Configurations}
\label{app:config}

\subsection{I2V Generation Parameters}
Table~\ref{tab:genparams} lists the exact inference configuration for every I2V model in IPV-Bench. We deliberately run each model at a demanding, near-default generation setting (up to $49$ frames and $\sim$6\,s of video), which is substantially more challenging than the short $\sim$1\,s clips ($25$ frames) that prior protection studies validate on.

\begin{table}[h]
\centering
\small
\setlength{\tabcolsep}{5pt}
\renewcommand{\arraystretch}{1.2}
\caption{Per-model I2V generation configuration. Duration is computed as frames\,/\,fps.}
\label{tab:genparams}
\begin{tabular}{llccccccc}
\toprule
Model & Checkpoint & Frames & Resolution & Steps & CFG & FPS & Duration & Access \\
\midrule
Wan2.2-TI2V-5B    & Wan2.2-TI2V-5B                       & 41 & $704{\times}1280$ & 30 & 5.0 & 16 & 2.6\,s & open \\
SkyReels-V2-I2V   & SkyReels-V2-I2V-1.3B-540P            & 49 & $544{\times}960$  & 30 & 6.0 & 8  & 6.1\,s & open \\
SVD-XT            & stable-video-diffusion-img2vid-xt    & 25 & $576{\times}1024$ & 30 & 1.0 & 8  & 3.1\,s & open \\
Seedance 1.5 Pro  & doubao-seedance-1-5-pro (API)        & 97 & $480{\times}848$  & -- & --  & 24 & 4.0\,s & black-box \\
\bottomrule
\end{tabular}

\vspace{2pt}
{\footnotesize SVD-XT decodes in chunks of 8 frames.}
\end{table}

\subsection{Protection Method Configurations}
All five protection methods are unified to a common $\ell_\infty$ budget of $\epsilon = 16/255 \approx 0.063$, so that measured differences reflect how a method \emph{allocates} its perturbation rather than its magnitude. Method-specific optimization settings follow each method's official implementation:

\begin{itemize}\itemsep1pt \topsep2pt
  \item \textbf{PhotoGuard}~\cite{salman2023photoguard}: encoder-latent attack, $200$ PGD steps with a decaying step size, $\ell_\infty$ projection at $\epsilon=16/255$.
  \item \textbf{Mist}~\cite{liang2023mist}: fused textural + semantic loss, $40$ PGD iterations, step size $0.01$, $\epsilon=16/255$.
  \item \textbf{I2VGuard}~\cite{i2vguard}: temporal-attention perturbation, $100$ Adam steps, learning rate $5{\times}10^{-3}$, $\alpha=2.0$, $\epsilon=16/255$.
  \item \textbf{VGMShield}~\cite{pang2025vgmshieldmitigatingmisusevideo}: feature-space PGD against the conditioning embedder, $1000$ diffusion steps, step size $\alpha=1/255$, $25$ denoising steps per iteration, $\epsilon=16/255$.
  \item \textbf{EditShield}~\cite{chen2024editshieldprotectingunauthorizedimage}: expectation-over-transformation optimization, $30$ steps, $\epsilon=16/255$.
  \item \textbf{RandomNoise} (reference): i.i.d.\ uniform noise sampled at the same $\ell_\infty$ budget, anchoring how much disruption is attributable to arbitrary perturbation.
\end{itemize}

\subsection{Preprocessing Attack Configurations}
The robustness study applies three lightweight, rule-based transforms that mirror the everyday operations images undergo on real sharing platforms:
\begin{itemize}\itemsep1pt \topsep2pt
  \item \textbf{JPEG compression} at quality factor $75$ (moderate, platform-typical compression).
  \item \textbf{Gaussian noise} with standard deviation $16/255 \approx 0.063$ (in $[0,1]$ pixel scale).
  \item \textbf{H.264 re-encode} via \texttt{libx264} at constant rate factor (CRF) $32$, matching the codec path a video-sharing pipeline would apply.
\end{itemize}

\section{Full Per-Domain Results}
\label{app:domain}
This section reports the complete per-domain breakdown behind Finding~2 and the aggregated main-paper tables. All numbers use the full IPV-500 set ($100$ images per domain) under the unified $\epsilon{=}16/255$ budget, over the four I2V models (Wan, SkyReel, SVD, Seedance). Table~\ref{tab:app_domain_fid} gives per-domain image fidelity (model-independent), Table~\ref{tab:app_domain_eff} per-domain effectiveness averaged over models, Table~\ref{tab:app_domain_vbench} the per-domain VBench degradation resolved into its six dimensions, and Table~\ref{tab:app_vbench_full} the six-dimension VBench breakdown per method and model that underlies the scale-normalized VB column of Table~\ref{tab:eff}.

% Auto-generated by gen_appendix_domain_tables.py -- do not edit by hand.
\begin{table}[H]
\centering\small
\setlength{\tabcolsep}{4pt}
\caption{\textbf{Per-domain image fidelity} (model-independent, $\epsilon{=}16/255$). Each block is one metric; columns are the five IPV-500 domains.}
\label{tab:app_domain_fid}
\begin{tabular}{lccccc}
\toprule
Method & Faces & Artworks & Nat.\ scenes & Objects & Animals \\
\midrule
\multicolumn{6}{l}{\emph{PSNR\,$\uparrow$}} \\
PhotoGuard & 32.0 & 31.8 & 32.0 & 31.9 & 31.9 \\
Mist & 27.9 & 27.0 & 27.1 & 26.1 & 27.6 \\
I2VGuard & 36.6 & 34.3 & 34.6 & 33.8 & 36.0 \\
VGMShield & 32.0 & 31.8 & 32.1 & 31.1 & 31.9 \\
EditShield & 34.0 & 34.4 & 34.4 & 34.4 & 34.2 \\
\midrule
\multicolumn{6}{l}{\emph{SSIM\,$\uparrow$}} \\
PhotoGuard & 0.797 & 0.843 & 0.803 & 0.826 & 0.824 \\
Mist & 0.708 & 0.727 & 0.689 & 0.706 & 0.732 \\
I2VGuard & 0.934 & 0.926 & 0.922 & 0.926 & 0.938 \\
VGMShield & 0.788 & 0.839 & 0.798 & 0.819 & 0.817 \\
EditShield & 0.846 & 0.900 & 0.855 & 0.877 & 0.874 \\
\midrule
\multicolumn{6}{l}{\emph{LPIPS\,$\downarrow$}} \\
PhotoGuard & 0.209 & 0.149 & 0.251 & 0.187 & 0.204 \\
Mist & 0.345 & 0.323 & 0.418 & 0.352 & 0.351 \\
I2VGuard & 0.071 & 0.086 & 0.132 & 0.100 & 0.076 \\
VGMShield & 0.257 & 0.178 & 0.301 & 0.222 & 0.256 \\
EditShield & 0.204 & 0.127 & 0.227 & 0.161 & 0.180 \\
\midrule
\multicolumn{6}{l}{\emph{CLIP-I\,$\uparrow$}} \\
PhotoGuard & 0.967 & 0.965 & 0.932 & 0.939 & 0.949 \\
Mist & 0.902 & 0.929 & 0.863 & 0.880 & 0.913 \\
I2VGuard & 0.977 & 0.986 & 0.962 & 0.947 & 0.960 \\
VGMShield & 0.931 & 0.958 & 0.905 & 0.915 & 0.935 \\
EditShield & 0.972 & 0.981 & 0.949 & 0.950 & 0.953 \\
\bottomrule
\end{tabular}
\end{table}

\begin{table}[H]
\centering\small
\setlength{\tabcolsep}{4pt}
\caption{\textbf{Per-domain protection effectiveness}, averaged over the four I2V models. CLIP-T degradation rate (\%, higher $=$ stronger) uses the paper's $(S_{\text{orig}}-S_{\text{prot}})/S_{\text{orig}}$ definition. FVD, being a distributional metric, does \emph{not} aggregate to the 500-image FVD and is confounded by intrinsic domain diversity (\S Finding 2); it is listed per domain for completeness only.}
\label{tab:app_domain_eff}
\begin{tabular}{lccccc}
\toprule
Method & Faces & Artworks & Nat.\ scenes & Objects & Animals \\
\midrule
\multicolumn{6}{l}{\emph{CLIP-T degradation rate (\%)}} \\
PhotoGuard & +0.63 & +0.05 & -0.65 & -0.58 & +0.49 \\
Mist & +1.22 & +1.55 & +1.77 & +0.04 & +1.17 \\
I2VGuard & +1.39 & +1.18 & +0.85 & +0.24 & +1.39 \\
VGMShield & +0.97 & +0.53 & +0.02 & -0.43 & +0.82 \\
EditShield & +0.70 & -0.55 & -0.12 & -0.54 & -0.00 \\
\midrule
\multicolumn{6}{l}{\emph{FVD}} \\
PhotoGuard & 6.3 & 9.9 & 5.9 & 12.4 & 9.2 \\
Mist & 9.9 & 13.5 & 9.8 & 16.2 & 15.9 \\
I2VGuard & 6.8 & 11.3 & 6.3 & 13.6 & 11.8 \\
VGMShield & 7.1 & 10.5 & 7.1 & 14.3 & 11.0 \\
EditShield & 6.2 & 9.4 & 5.9 & 12.0 & 8.9 \\
\bottomrule
\end{tabular}
\end{table}

\begin{table}[H]
\centering\small
\setlength{\tabcolsep}{3.5pt}
\caption{\textbf{Per-domain VBench degradation by dimension} (\%, higher $=$ stronger protection), averaged over all methods and models. This is the dimension-level breakdown aggregated into the VB column of Table~\ref{tab:eff}.}
\label{tab:app_domain_vbench}
\begin{tabular}{lcccccc}
\toprule
Domain & Subj.\ cons. & Bkg.\ cons. & Motion & Aesth. & Imaging & Temp.\ flk. \\
\midrule
Faces & +2.93 & +2.09 & +0.28 & +2.86 & +5.30 & +0.21 \\
Artworks & +3.11 & +1.52 & +0.33 & +6.96 & +7.65 & +0.27 \\
Nat.\ scenes & +2.29 & +1.55 & +0.15 & +9.56 & +8.01 & +0.02 \\
Objects & +1.75 & +1.73 & +0.02 & +4.27 & +7.17 & -0.02 \\
Animals & +2.33 & +1.49 & +0.24 & +3.60 & +4.49 & +0.16 \\
\bottomrule
\end{tabular}
\end{table}

\begin{table*}[t]
\centering\small
\setlength{\tabcolsep}{4pt}
\caption{\textbf{VBench degradation by dimension, per method and model} (\%, higher $=$ stronger protection; full 500-image set). This un-aggregates the scale-normalized VB column of Table~\ref{tab:eff} into its six raw VBench dimensions.}
\label{tab:app_vbench_full}
\begin{tabular}{llcccccc}
\toprule
Model & Method & Subj.\ cons. & Bkg.\ cons. & Motion & Aesth. & Imaging & Temp.\ flk. \\
\midrule
Wan & PhotoGuard & +1.02 & +0.36 & +0.14 & +7.05 & +7.88 & +0.06 \\
 & Mist & +7.21 & +3.60 & +1.45 & +17.53 & +20.56 & +2.20 \\
 & I2VGuard & +4.36 & +3.04 & +1.18 & +19.91 & +7.79 & +2.10 \\
 & VGMShield & +0.88 & +0.14 & +0.10 & +8.60 & +9.65 & +0.03 \\
 & EditShield & -0.06 & -0.17 & -0.06 & +3.30 & +3.38 & -0.25 \\
\midrule
SkyReel & PhotoGuard & +1.74 & +1.89 & -0.08 & +2.00 & +3.23 & -0.29 \\
 & Mist & +4.37 & +2.33 & -0.04 & +5.70 & +7.13 & -0.21 \\
 & I2VGuard & +0.66 & +1.12 & -0.13 & +1.26 & +1.75 & -0.23 \\
 & VGMShield & +1.64 & +2.09 & -0.05 & +3.43 & +4.73 & -0.20 \\
 & EditShield & +0.54 & +1.14 & -0.05 & +2.93 & +2.00 & -0.22 \\
\midrule
SVD & PhotoGuard & +1.46 & +1.32 & +0.36 & +7.58 & +8.76 & -0.12 \\
 & Mist & +6.06 & +1.65 & +0.80 & +15.63 & +19.95 & -0.12 \\
 & I2VGuard & +0.89 & +0.62 & +0.08 & +4.41 & +9.19 & -0.21 \\
 & VGMShield & +3.20 & +1.60 & +0.02 & +9.51 & +10.21 & -0.18 \\
 & EditShield & +1.14 & +0.53 & +0.31 & +3.57 & +4.01 & -0.02 \\
\midrule
Seedance & PhotoGuard & +2.36 & +2.24 & +0.02 & -0.00 & +0.60 & +0.01 \\
 & Mist & +6.63 & +3.61 & +0.03 & -0.11 & +4.49 & +0.08 \\
 & I2VGuard & +0.85 & +1.07 & -0.00 & -0.31 & +2.93 & +0.00 \\
 & VGMShield & +3.07 & +3.25 & +0.03 & +0.08 & +0.86 & +0.07 \\
 & EditShield & +1.49 & +2.10 & +0.04 & -0.20 & +0.78 & +0.12 \\
\bottomrule
\end{tabular}
\end{table*}

\section{Licenses}
\label{app:licenses}
Table~\ref{tab:licenses} lists the licenses of all assets used in IPV-Bench. IPV-500 redistributes only image identifiers and derived prompts where source licenses restrict image redistribution; all source images are obtained under their original terms. We will release the IPV-Bench evaluation framework and IPV-500 metadata under a permissive research license upon publication.

\begin{table}[h]
\centering
\small
\setlength{\tabcolsep}{6pt}
\renewcommand{\arraystretch}{1.2}
\caption{Licenses of datasets, models, and method implementations used in IPV-Bench.}
\label{tab:licenses}
\begin{tabular}{lll}
\toprule
Asset & Type & License \\
\midrule
FFHQ            & Dataset & Creative Commons BY-NC-SA / public-domain subsets \\
WikiArt         & Dataset & Research use (per-artwork terms) \\
LHQ             & Dataset & Research use \\
MS-COCO         & Dataset & Creative Commons BY 4.0 \\
AFHQ-v2         & Dataset & Creative Commons BY-NC 4.0 \\
\midrule
Qwen3-VL / Qwen2.5-VL & Model & Apache-2.0 \\
InternVL2       & Model & MIT \\
Wan2.2-TI2V-5B  & Model & Apache-2.0 \\
SkyReels-V2-I2V & Model & Apache-2.0 \\
Stable Video Diffusion & Model & Stability AI Community License \\
Seedance 1.5 Pro & API   & Commercial API terms of service \\
\midrule
PhotoGuard, Mist, EditShield & Code & MIT / research use \\
I2VGuard, VGMShield          & Code & Research use (per repository) \\
\bottomrule
\end{tabular}
\end{table}

\section{Ethics Statement}
\label{app:ethics}

\paragraph{Purpose and framing.}
IPV-Bench is a \emph{defensive} evaluation effort. Our goal is to measure, under a fair and unified protocol, how well existing image-protection methods actually defend against image-to-video misuse. The study develops no new attack: every protection method and every generator we evaluate is already publicly available, and our preprocessing transforms (JPEG, Gaussian noise, H.264) are standard, non-adversarial operations that any image inevitably undergoes on sharing platforms. We introduce no novel capability for producing harmful content.

\paragraph{Dual-use considerations.}
Our central findings are negative: current perturbation-based protections are weak, do not transfer across generators, and are easily neutralized by routine preprocessing. Reporting such weaknesses carries a dual-use tension, since an adversary could read them as confirmation that today's defenses are easy to bypass. We judge the disclosure to be net-beneficial, for the same reason robustness benchmarks in adversarial ML and watermarking are published: users and platforms currently \emph{overestimate} the protection these methods provide, and quantifying the gap is a precondition for building defenses that genuinely work. We frame every finding as motivation for stronger, frequency-aware and multi-surrogate defenses rather than as an exploitation recipe.

\paragraph{Data and privacy.}
All images originate from established, publicly released research datasets used under their original licenses (Table~\ref{tab:licenses}). We do not scrape new personal data, do not target or identify real individuals, and do not release any generated video as standalone content. Face images come from FFHQ, a dataset already curated for research use; we use them solely to measure protection fidelity and disruption, not to build recognition or profiling systems. Generated videos exist only as intermediate artifacts for computing evaluation metrics and are not distributed as deepfakes.

\paragraph{Potential harms and mitigations.}
The primary residual risk is that our released benchmark lowers the barrier to evaluating, and thus circumventing, protection methods. We mitigate this by releasing evaluation code and dataset \emph{metadata} rather than a turnkey misuse system, and by centering the release on defense development. No human-subjects experiments were conducted, and the work required no IRB review.

% ---------------------------------------------------------------
% ---------------------------------------------------------------

\end{document}